\documentclass{article}

\usepackage[dblblindworkshop, final]{neurips_2025}

\usepackage[utf8]{inputenc} % allow utf-8 input
\usepackage[T1]{fontenc}    % use 8-bit T1 fonts
\usepackage{hyperref}       % hyperlinks
\usepackage{url}            % simple URL typesetting
\usepackage{booktabs}       % professional-quality tables
\usepackage{amsfonts}       % blackboard math symbols
\usepackage{nicefrac}       % compact symbols for 1/2, etc.
\usepackage{microtype}      % microtypography
\usepackage{xcolor}         % colors
\usepackage{amsmath}
\usepackage{graphicx}
\usepackage{subcaption}
\usepackage{multirow} 

\title{Learning to Align Molecules and Proteins: A Geometry-Aware Approach to Binding Affinity}

\author{%
  Mohammadsaleh Refahi \\
  Drexel University \\
  Philadelphia, PA \\
  \texttt{sr3622@drexel.edu} \\
  \And
  Bahrad A. Sokhansanj \\
  Drexel University \\
  Philadelphia, PA \\
  \texttt{bahrad@molhealtheng.com} \\
  \And
  James R. Brown \\
  Drexel University \\
  Philadelphia, PA \\
  \texttt{jb4633@drexel.edu} \\
  \And
  Gail Rosen \\
  Drexel University \\
  Philadelphia, PA\\
  \texttt{glr26@drexel.edu} \\
}

\begin{document}

\maketitle

\begin{abstract}
Accurate prediction of drug–target binding affinity can accelerate drug discovery by prioritizing promising compounds before costly wet-lab screening. While deep learning has advanced this task, most models fuse ligand and protein representations via simple concatenation and lack explicit geometric regularization, resulting in poor generalization across chemical space and time. We introduce \textbf{FIRM-DTI}, a lightweight framework that conditions molecular embeddings on protein embeddings through a feature-wise linear modulation (FiLM) layer and enforces metric structure with a triplet loss. An RBF regression head operating on embedding distances yields smooth, interpretable affinity predictions. Despite its modest size, {FIRM-DTI} achieves state-of-the-art performance on the Therapeutics Data Commons DTI-DG benchmark, as demonstrated by an extensive ablation study and out-of-domain evaluation. Our results underscore the value of conditioning and metric learning for robust drug–target affinity prediction.(Code available at \url{https://github.com/EESI/Firm-DTI})
\end{abstract}

\section{Introduction}

The process of discovering a new therapeutic agent hinges on understanding how strongly a candidate molecule binds to a protein target.  Binding affinity determines pharmacological potency and selectivity, but measuring it experimentally is laborious and expensive.  Computational predictions of binding affinity therefore play an important role in virtual screening and lead optimization.  Traditional quantitative structure–activity relationship models rely on handcrafted descriptors and are often restricted to narrow chemical series.  Deep neural networks promise to learn richer representations from raw sequences and graphs, yet many state‑of‑the‑art methods treat the two modalities independently and use ad‑hoc interaction layers without explicit regularization.  Furthermore, high‑quality affinity data are scarce relative to the vast chemical and proteomic space, making generalization to novel drugs and targets challenging.

This work proposes a simple and effective architecture that addresses these limitations.  Instead of concatenating ligand and protein features, we employ a FiLM conditioning layer to modulate molecular embeddings by protein embeddings, allowing the model to learn target‑specific transformations.  We additionally incorporate a triplet metric‑learning objective that pulls interacting pairs together and pushes non‑interacting pairs apart in the embedding space.  A radial basis function (RBF) regression head then maps distances to continuous affinity values.  As we show on a temporally split benchmark, this combination yields competitive performance with far fewer parameters than recent large models.

\section{Background}

Deep learning has become a cornerstone of drug--target interaction (DTI) and binding affinity prediction.  Early sequence-based methods typically embedded drug SMILES strings and protein sequences with recurrent or convolutional networks, then combined the embeddings with a feed-forward layer.  More recent representation learning approaches have introduced graph neural networks for molecular structures and large protein language models for amino acid sequences.  For instance, MolE employs a disentangled attention transformer to produce atom-level graph embeddings \cite{mendez2024mole}, while ESM leverages masked language modeling over massive protein corpora to yield residue-aware embeddings \cite{lin2023evolutionary}.  ChemBERTa\cite{chithrananda2020chemberta} and rxnfp\cite{schwaller2021mapping} have further demonstrated the utility of pretrained transformers directly on SMILES strings for chemical property prediction.

Hybrid architectures often concatenate drug and protein embeddings and pass them through a multilayer perceptron, but this strategy ignores conditional dependencies between ligands and targets\cite{thafar2022affinity2vec}.  Knowledge-based models attempt to mitigate this by integrating curated interaction networks or graph-derived embeddings (e.g., node2vec over protein--protein interaction graphs), yet such external information is costly to maintain and may not generalize across chemical space\cite{lam2023otter,fattahi2019drug}.

Recent studies show that residue-level protein language model embeddings contain rich contextual signals for drug–target interaction tasks, especially when combined with transfer learning strategies such as contrastive or task-specific pretraining \cite{naderializadeh2025aggregating,singh2023contrastive,sledzieski2022adapting}.  However, most sequence-based DTI models still rely on simple concatenation of drug and protein representations, which limits their ability to capture robust geometric relationships in the latent space.  Explicit geometry-aware designs remain underexplored: while protein PLMs enable similarity learning at the residue level, few models have incorporated metric-based objectives to enforce clustering of interacting pairs.

\section{Methods}

\begin{figure}[htbp]
  \centering
  \includegraphics[width=0.96\textwidth]{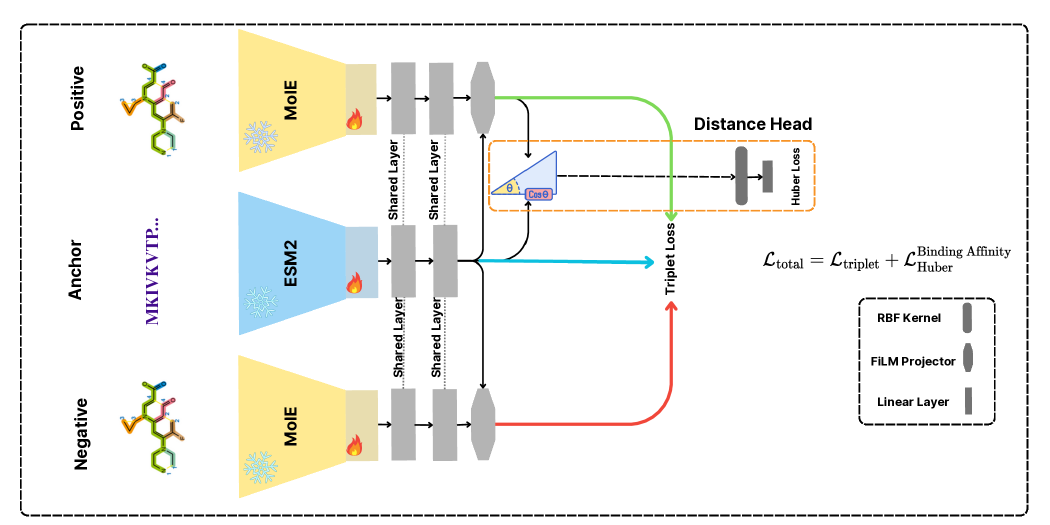}
    \caption{FIRM-DTI architecture. Drug embeddings (from MolE) and protein embeddings (from ESM2) are projected into a shared latent space via a FiLM conditioning layer. The model is trained with a triplet loss to enforce geometric alignment between interacting pairs, and an RBF-based prediction head is used to regress binding affinity values.}
  \label{fig:architecture}
\end{figure}

Our framework consists of three stages: featurization of molecules and proteins, conditioning via a FiLM layer, and distance‑based affinity regression.  Figure~\ref{fig:architecture} summarizes the architecture.

\subsection{Feature extraction}

We obtain latent representations for ligands and proteins from pretrained experts.  The MolE model treats a molecule as a graph where nodes represent atoms and edges represent bonds.  Atom features include Daylight atomic invariants and the Morgan fingerprint, and positional information is encoded through relative bond distances.  A disentangled attention transformer then produces an embedding $z_d \in \mathbb{R}^d$ for each molecule \cite{mendez2024mole}.  For the target protein, we use the ESM2 language model which processes amino acid sequences and outputs a per‑sequence embedding $z_t \in \mathbb{R}^d$ by averaging residue representations \cite{lin2023evolutionary}.

\subsection{FiLM conditioning}

To incorporate target context, we apply a FiLM layer \cite{perez2018film}.  Given drug embedding $z_d$ and protein embedding $z_t$, the conditioned embedding is
\[
\mathrm{FiLM}(z_d \mid z_t) = \gamma(z_t) \odot z_d + \beta(z_t),
\]
where $\gamma$ and $\beta$ are learned linear functions of $z_t$ and $\odot$ denotes element‑wise multiplication.  This transformation allows the model to scale and shift molecular features based on the target, capturing conditional interactions more flexibly than concatenation.

\subsection{Distance‑based prediction head}

After conditioning, we normalize the drug and protein embeddings and compute their cosine distance:
\[
\mathrm{dist}(\tilde{z}_d, \tilde{z}_t) = 1 - \frac{\tilde{z}_d \cdot \tilde{z}_t}{\|\tilde{z}_d\|\, \|\tilde{z}_t\|}.
\]
This distance is passed through a set of radial basis functions with $k$ centers $\mu_j$ evenly spaced in $[0,2]$:
\[
\phi_j = \exp\bigl(-((\mathrm{dist}(\tilde{z}_d,\tilde{z}_t) - \mu_j)^2)/(2\sigma^2)\bigr), \quad j=1,\dots,k.
\]
The final affinity prediction is computed via a linear layer $y_{\mathrm{pred}} = W \phi + b$.  This head enforces that similar embeddings yield similar predictions and provides a smooth mapping from distances to continuous affinities.

\subsection{Training objective}

Training requires both positive and negative drug–protein pairs.  Positive examples are experimentally validated interactions with known binding affinities, while negative examples are randomly paired molecules and targets with no reported interaction.  During training, we minimize a combination of a triplet loss and a Huber regression loss:
\[
\mathcal{L}_{\text{total}} = \mathcal{L}_{\text{triplet}} + \mathcal{L}_{\text{Huber}}^{\text{affinity}},
\]
where
\[
\mathcal{L}_{\text{triplet}} = \max\bigl(0, d(f(x_a),f(x_p)) - d(f(x_a),f(x_n)) + \alpha\bigr)
\]
encourages the distance between an anchor $x_a$ and a positive $x_p$ to be smaller than the distance to a negative $x_n$ by margin $\alpha$, and the Huber loss with threshold $\delta$ stabilizes regression:
\[
\mathcal{L}_{\text{Huber}}^{\text{affinity}} =
\begin{cases}
\tfrac{1}{2}(y - \hat{y})^2, & \text{if } |y - \hat{y}| \leq \delta,\\
\delta\,|y - \hat{y}| - \tfrac{1}{2}\delta^2, & \text{otherwise}.
\end{cases}
\]

\subsection{Dataset and evaluation protocol}

We evaluate on the Drug–Target Interaction Domain Generalization (DTI‑DG) benchmark introduced by the Therapeutics Data Commons \cite{huang2021therapeutics}.  The benchmark partitions binding affinity data derived from BindingDB by patent year, training models on interactions from 2013–2018 and testing on interactions from 2019–2021.  This temporal split mimics the challenge of predicting affinities for newly patented drugs and targets.  All models are trained for 25 epochs with identical hyperparameters, and performance is measured by Pearson correlation coefficient (PCC) between predicted and true affinities.

\paragraph{DTI-Datasets}
We evaluate on three standard DTI benchmarks: \textbf{DAVIS} \cite{davis2011comprehensive}, \textbf{BindingDB} \cite{liu2007bindingdb}, and \textbf{BIOSNAP} (ChG-Miner) \cite{zitnik2018biosnap}. 
Following \textbf{MolTrans} \cite{huang2021moltrans}, we binarize DAVIS and BindingDB using $K_d<30$ (positive) and $K_d\ge 30$ (negative). 
BIOSNAP contains only positives; consistent with MolTrans, we generate negatives by randomly pairing proteins and compounds in equal number to the positives. Each dataset is split into 70\% train, 10\% validation, and 20\% test.

\section{Results}

\subsection{Ablation study}

To understand the contributions of each component, we conduct an ablation study on DTI‑DG.  Table~\ref{tab:ablation} reports the mean PCC over the test domains when removing the FiLM conditioning layer or the triplet loss.  Eliminating the FiLM projector leads to a modest decline in performance, while omitting the triplet loss causes a severe drop, highlighting the importance of metric learning.

\begin{table}[t]
  \centering
  \caption{Impact of model components on DTI‑DG performance. }
  \label{tab:ablation}
  \begin{tabular}{lc}
    \toprule
    \textbf{Model variant} & \textbf{PCC}\,\(\uparrow\) \\
    \midrule
    Full model & 0.59 \\
    \hspace{1em}-- without FiLM conditioning  & 0.55 \\
    \hspace{1em}-- without triplet loss  & 0.32 \\
    \bottomrule
  \end{tabular}
\end{table}

\subsection{Out-of-domain prediction}
\begin{figure}[t]
  \centering
  \begin{subfigure}[t]{0.49\linewidth}
    \centering
    \includegraphics[width=\linewidth]{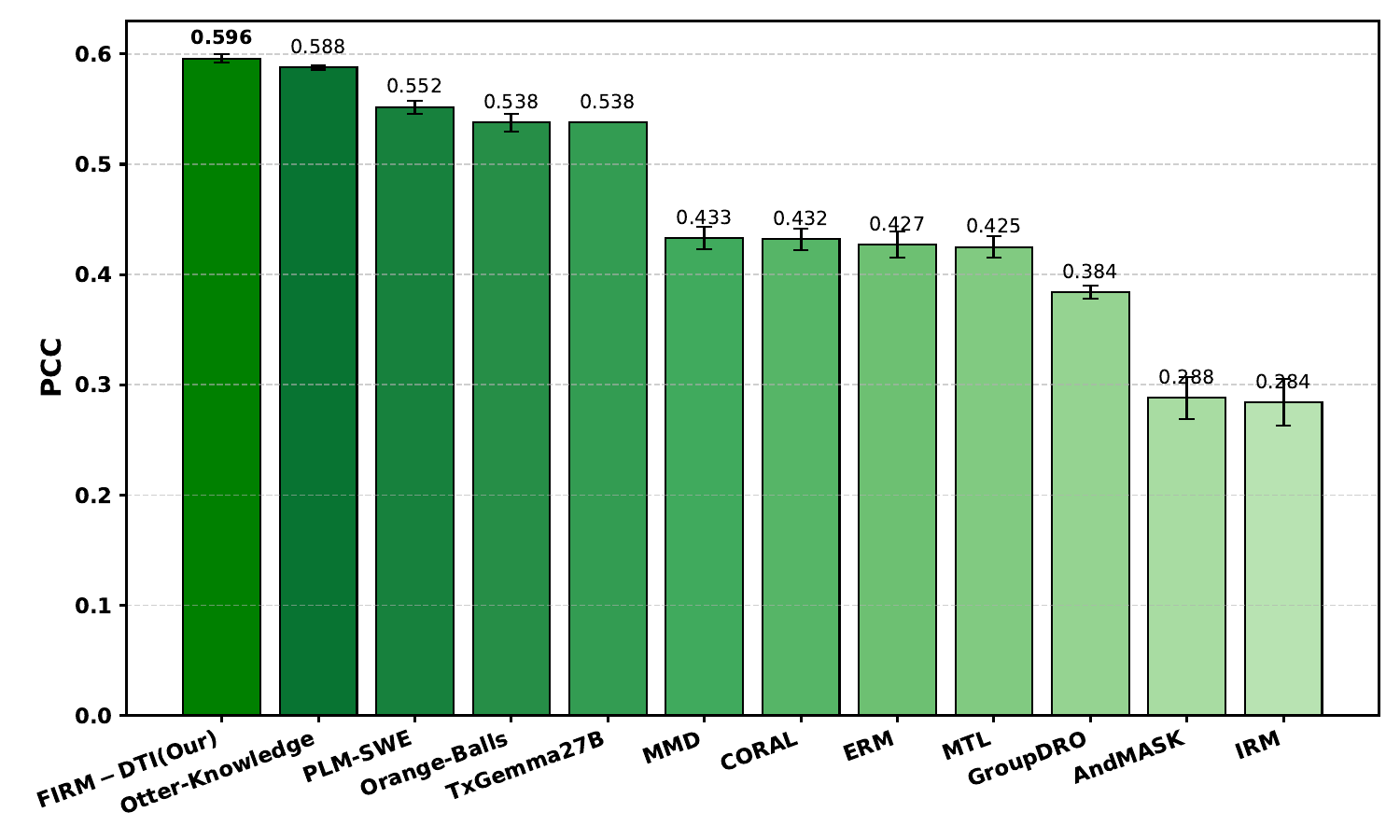}
    \caption{PCC on DTI-DG benchmark.}
  \end{subfigure}\hfill
  \begin{subfigure}[t]{0.44\linewidth}
    \centering
    \includegraphics[width=\linewidth]{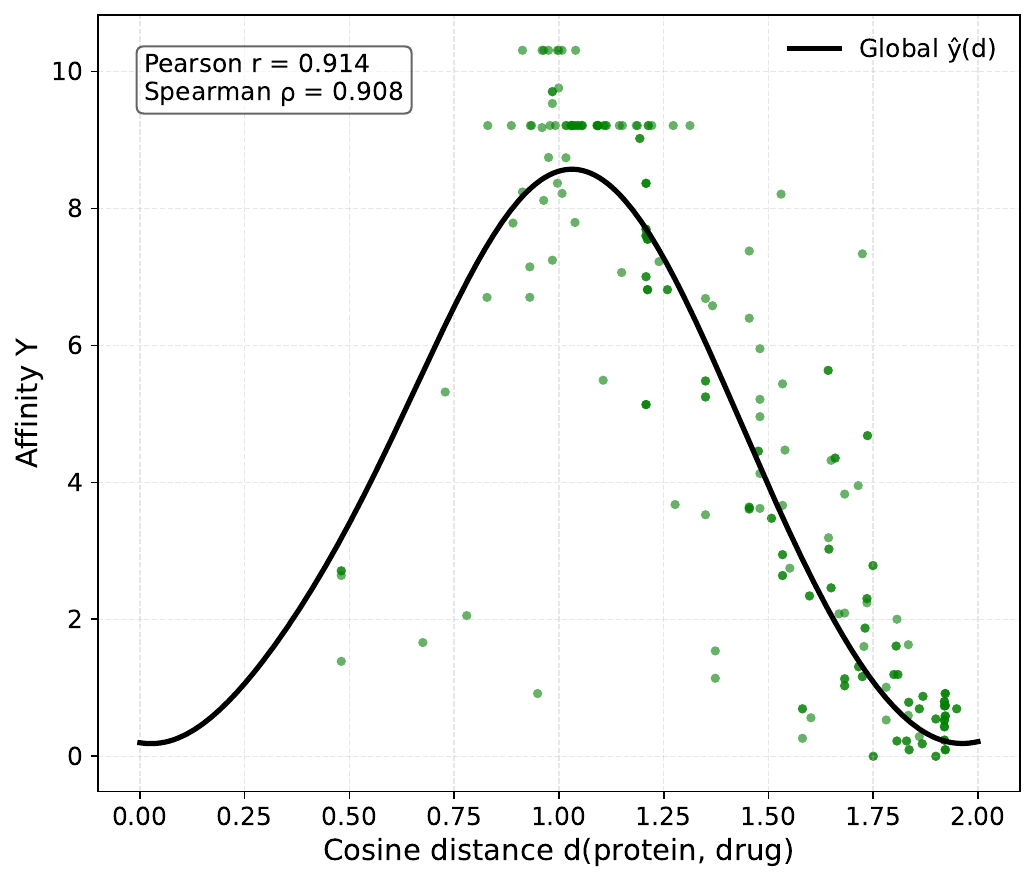}
    \caption{Distance–affinity mapping.}
  \end{subfigure}
  \vspace{-0.5em}
  \caption{
  (a) Our model outperforms recent baselines on DTI-DG despite fewer parameters and no external knowledge. 
  (b) The learned RBF head produces a smooth mapping $\hat{y}(d)$ from cosine distance to affinity ($r{=}0.91$), demonstrating geometry-awareness.
  }
  \label{fig:ood}
  \vspace{-0.5em}
\end{figure}

Figure~\ref{fig:ood} compares our method against recent baselines on the DTI-DG test set. Despite using fewer parameters and no external knowledge (unlike the Otter-Knowledge ensemble~\cite{lam2023otter}), our model achieves a PCC of 0.59. This performance surpasses state-of-the-art sequence-based models such as PLM-SWE~\cite{naderializadeh2025aggregating} and TxGemma27B~\cite{wang2025txgemma}, as well as network-based approaches like Otter-Knowledge~\cite{lam2023otter}. These results highlight that while some models rely on massive scale (e.g., 27B parameters in TxGemma) or external knowledge (e.g., Otter-Knowledge), our approach attains superior performance through conditioning and metric learning alone. As shown in Fig.~\ref{fig:ood}b, the learned RBF head with $\sigma=0.2$ produces a smooth mapping $\hat{y}(d)=\sum_j w_j \exp\!\left(-\tfrac{(d-\mu_j)^2}{2\sigma^2}\right)$ from cosine distance $d$ to affinity, yielding $r=0.91$ correlation and confirming that, unlike baselines in Fig.~\ref{fig:ood}a, our model encodes binding affinity directly as a function of embedding geometry.

\subsection{DTI Experiments}
\label{sec:dti}

In addition to binding–affinity regression, we trained \emph{drug–target interaction} (DTI) classifiers. Our setup follows \cite{sledzieski2022adapting,singh2023contrastive} with two modifications. First, we replace the Huber regression objective with \texttt{BCEWithLogitsLoss} to model binary interaction labels directly. Second, we adopt a contrastive (triplet-style) sampling strategy for continual learning: within each mini-batch the \emph{anchor} is the protein (target), the \emph{positive} is a drug known to interact with that protein, and the \emph{negative} is a different drug not known to interact with that protein (negatives can be resampled each epoch).

Table~\ref{tab:plm_vs_nonplm} compares FIRM-DTI with strong baselines on three standard DTI datasets. 
Across BIOSNAP and BindingDB our model consistently achieves the highest or comparable AUPR and AUROC scores, 
showing that the FiLM-conditioned, geometry-aware representation generalizes well beyond affinity regression. 
On the much smaller and more imbalanced DAVIS set, performance drops for all methods but FIRM-DTI remains 
competitive, underscoring its robustness under challenging data distributions.

\begin{table*}[t]
\centering
\caption{Comparison on BIOSNAP, BindingDB, and DAVIS. Mean $\pm$ s.e.m. over 5 random seeds. 
 \cite{sledzieski2022adapting}.}
\label{tab:plm_vs_nonplm}
\begin{tabular}{l lcc}
\toprule
\textbf{Benchmark} & \textbf{Model} & \textbf{AUPR} & \textbf{AUROC} \\
\midrule
\multirow{6}{*}{\textbf{BIOSNAP}} 
& \textbf{FIRM-DTI (Our Result)} & \textbf{0.919 $\pm$ 0.016} & \textbf{0.910 $\pm$ 0.004} \\
& ConPLex~\cite{singh2023contrastive} & $0.897 \pm 0.001$ & $-$ \\
& ProtBert + Morgan~\cite{sledzieski2022adapting} & $0.895 \pm 0.004$ & $0.873 \pm 0.004$ \\
& MolTrans~\cite{huang2021moltrans} & $0.885 \pm 0.005$ & $0.876 \pm 0.007$ \\
& GNN-CPI~\cite{tsubaki2019compound} & $0.890 \pm 0.004$ & $0.879 \pm 0.007$ \\
& DeepConv-DTI~\cite{lee2019deepconv} & $0.889 \pm 0.005$ & $0.883 \pm 0.002$ \\
\midrule
\multirow{6}{*}{\textbf{BindingDB}} 
& \textbf{FIRM-DTI (Our Result)} & $0.647 \pm 0.003$ & \textbf{0.916 $\pm$ 0.001} \\
& ConPLex~\cite{singh2023contrastive} & $0.628 \pm 0.012$ & $-$ \\
& ProtBert + Morgan~\cite{sledzieski2022adapting} & \textbf{0.652 $\pm$ 0.005} & $0.876 \pm 0.007$ \\
& MolTrans~\cite{huang2021moltrans} & $0.598 \pm 0.013$ & $0.898 \pm 0.009$ \\
& GNN-CPI~\cite{tsubaki2019compound} & $0.578 \pm 0.015$ & $0.900 \pm 0.004$ \\
& DeepConv-DTI~\cite{lee2019deepconv} & $0.611 \pm 0.015$ & $0.908 \pm 0.004$ \\
\midrule
\multirow{6}{*}{\textbf{DAVIS}} 
& \textbf{FIRM-DTI (Our Result)} & $0.460 \pm 0.004$ & $0.880 \pm 0.001$ \\
& ConPLex~\cite{singh2023contrastive} & $0.458 \pm 0.016$ & $-$ \\
& ProtBert+Morgan~\cite{sledzieski2022adapting} & \textbf{0.511 $\pm$ 0.012} & \textbf{0.917 $\pm$ 0.003} \\
& MolTrans~\cite{huang2021moltrans} & $0.335 \pm 0.017$ & $0.889 \pm 0.007$ \\
& GNN-CPI~\cite{tsubaki2019compound} & $0.269 \pm 0.020$ & $0.840 \pm 0.012$ \\
& DeepConv-DTI~\cite{lee2019deepconv} & $0.299 \pm 0.039$ & $0.884 \pm 0.008$ \\
\bottomrule
\end{tabular}
\end{table*}

\section{Discussion}

Our study shows that a relatively small model can achieve competitive accuracy on challenging drug–target affinity prediction tasks when equipped with the right inductive biases.  The FiLM conditioning layer allows the network to learn target‑specific transformations of molecular features, and the triplet loss enforces a geometric structure that aligns interacting pairs.  Together, these components improve out‑of‑domain generalization without relying on large architectures or external knowledge bases.  A limitation of our work is the reliance on pretrained experts; future research may explore joint training of molecular and protein embeddings or incorporate three‑dimensional structural information.  Additionally, investigating uncertainty estimation could further accelerate practical drug discovery workflows.

\bibliographystyle{plainnat}
\bibliography{ref,extra}

\section*{Technical Appendices}

\subsection*{Training details}
We trained our model on a single NVIDIA H100 GPU with 80~GiB of memory.  The total number of learnable parameters is \(
\approx 1.2\times10^8\), of which the vast majority reside in the pretrained encoders.  For the protein encoder we use the ESM2 model \texttt{esm2\_t12\_35M\_UR50D} from Facebook AI's repository\footnote{\url{https://huggingface.co/facebook/esm2\_t12\_35M\_UR50D}}; the molecular encoder is MolE, trained on the GuacaMol dataset\footnote{\url{https://codeocean.com/capsule/2105466/tree/v1}}.  Protein sequences are truncated or padded to a maximum length of 1{,}500 residues.

We optimize with the AdamW algorithm (\(\epsilon=10^{-6}\)) and use a cosine learning rate schedule with linear warmup.  The main hyperparameters are summarized in Table~\ref{tab:training}.

The sensitivity of performance to the Huber loss threshold $\delta$ is shown in Table~\ref{tab:delta}.

\begin{table}[h]
  \centering
  \caption{Effect of Huber loss threshold $\delta$ on PCC (DTI-DG).}
  \label{tab:delta}
  \begin{tabular}{cc}
    \toprule
    \textbf{$\delta$} & \textbf{PCC} \\
    \midrule
    0.75 & 0.56 \\
    0.50 & 0.59 \\
    0.25 & 0.58 \\
    \bottomrule
  \end{tabular}
\end{table}

\begin{table}[h]
  \centering
  \caption{Training hyperparameters.}
  \label{tab:training}
  \begin{tabular}{ll}
    \toprule
    \textbf{Hyperparameter} & \textbf{Value} \\
    \midrule
    Batch size & 24 drug--protein pairs \\
    Learning rate & $5\times10^{-5}$ (cosine decay, 500 warmup steps) \\
    Optimizer & AdamW, $\epsilon=10^{-6}$ \\
    Weight decay & 0.1 \\
    Triplet margin $\alpha$ & 0.9 \\
    Max protein length & 1500 residues \\
    Epochs & 25 \\
    Total parameters & $\approx 120M$ \\
    GPU used & NVIDIA H100 80~GiB \\
    Training time & $\sim$29 hours per run \\
    \bottomrule
  \end{tabular}
\end{table}

\newpage
\section*{NeurIPS Paper Checklist}

\begin{enumerate}

\item {\bf Claims}
    \item[] Question: Do the main claims made in the abstract and introduction accurately reflect the paper's contributions and scope?
    \item[] Answer: \answerYes{} % Replace by \answerYes{}, \answerNo{}, or \answerNA{}.
    \item[] Justification: We carefully reviewed the abstract and introduction, and they do not contain any claims that are not justified in the paper.

\item {\bf Limitations}
    \item[] Question: Does the paper discuss the limitations of the work performed by the authors?
    \item[] Answer: \answerYes{} % Replace by \answerYes{}, \answerNo{}, or \answerNA{}.
    \item[] Justification: We explicitly acknowledge limitations in Section~6, including reliance on pretrained experts, the absence of 3D structural information, and the need for future work on uncertainty estimation.

\item {\bf Theory assumptions and proofs}
    \item[] Question: For each theoretical result, does the paper provide the full set of assumptions and a complete (and correct) proof?
    \item[] Answer: \answerYes{} % Replace by \answerYes{}, \answerNo{}, or \answerNA{}.
    \item[] Justification: We provide all necessary assumptions and a complete derivation of the loss functions, including the triplet and Huber objectives.

    \item {\bf Experimental result reproducibility}
    \item[] Question: Does the paper fully disclose all the information needed to reproduce the main experimental results of the paper to the extent that it affects the main claims and/or conclusions of the paper (regardless of whether the code and data are provided or not)?
    \item[] Answer: \answerYes{} % Replace by \answerYes{}, \answerNo{}, or \answerNA{}.
    \item[] Justification:We use publicly available datasets and describe all model architectures, hyperparameters, and training procedures in detail. The code will be released upon acceptance to further support reproducibility.

\item {\bf Open access to data and code}
    \item[] Question: Does the paper provide open access to the data and code, with sufficient instructions to faithfully reproduce the main experimental results, as described in supplemental material?
    \item[] Answer: \answerYes{}
    \item[] Justification: All datasets used are publicly available and cited appropriately. We will release the code and detailed instructions for reproducing the main experimental results in the supplementary material and a public repository upon acceptance.

\item {\bf Experimental setting/details}
    \item[] Question: Does the paper specify all the training and test details (e.g., data splits, hyperparameters, how they were chosen, type of optimizer, etc.) necessary to understand the results?
     \item[] Answer: \answerYes{}
        \item[] Justification: The paper provides detailed descriptions of data splits, optimizer settings, training schedules, and hyperparameters in Appendix. These include the batch size, learning rate, number of epochs, and evaluation metrics used across all benchmarks.

\item {\bf Experiment statistical significance}
    \item[] Question: Does the paper report error bars suitably and correctly defined or other appropriate information about the statistical significance of the experiments?
    \item[] Answer: \answerYes{}
    \item[] Justification: For datasets with official splits, we follow the provided train/test partitions without additional resampling.

\item {\bf Experiments compute resources}
    \item[] Question: For each experiment, does the paper provide sufficient information on the computer resources (type of compute workers, memory, time of execution) needed to reproduce the experiments?
    \item[] Answer: \answerYes{}
    \item[] Justification: We report the GPU type, memory configuration, and training time estimates in  Appendix 6.1  These include hardware specifications such as H100 GPUs and per-task training durations to support reproducibility.

\item {\bf Code of ethics}
    \item[] Question: Does the research conducted in the paper conform, in every respect, with the NeurIPS Code of Ethics \url{https://neurips.cc/public/EthicsGuidelines}?
    \item[] Answer: \answerYes{} % Replace by \answerYes{}, \answerNo{}, or \answerNA{}.
    \item[] Justification: The research conducted in the paper conform, in every respect, with the
        NeurIPS Code of Ethics.

\item {\bf Broader impacts}
    \item[] Question: Does the paper discuss both potential positive societal impacts and negative societal impacts of the work performed?
    \item[] Answer: \answerYes{}
    \item[] Justification: We outline positive impacts for accelerating drug discovery and potential risks related to misuse or overreliance on computational predictions.

\item {\bf Safeguards}
    \item[] Question: Does the paper describe safeguards that have been put in place for responsible release of data or models that have a high risk for misuse (e.g., pretrained language models, image generators, or scraped datasets)?
    \item[] Answer: \answerNA{}
    \item[] Justification: The datasets and models used in this work are not considered high risk for misuse; they are based on publicly available genomic data and do not involve sensitive personal or dual-use information.

\item {\bf Licenses for existing assets}
    \item[] Question: Are the creators or original owners of assets (e.g., code, data, models), used in the paper, properly credited and are the license and terms of use explicitly mentioned and properly respected?
    \item[] Answer: \answerYes{}
    \item[] Justification: All external datasets and models used in this work are properly cited, and we rely exclusively on publicly released assets with appropriate licenses, as noted in the references and Appendix~A.

\item {\bf New assets}
    \item[] Question: Are new assets introduced in the paper well documented and is the documentation provided alongside the assets?
    \item[] Answer: \answerNA{}
    \item[] Justification: This work does not introduce any new datasets, models, or software assets beyond those already publicly available and cited.

\item {\bf Crowdsourcing and research with human subjects}
    \item[] Question: For crowdsourcing experiments and research with human subjects, does the paper include the full text of instructions given to participants and screenshots, if applicable, as well as details about compensation (if any)? 
    \item[] Answer: \answerNA{} % Replace by \answerYes{}, \answerNo{}, or \answerNA{}.
    \item[] Justification: This research does not involve crowdsourcing or human subjects.

\item {\bf Institutional review board (IRB) approvals or equivalent for research with human subjects}
    \item[] Question: Does the paper describe potential risks incurred by study participants, whether such risks were disclosed to the subjects, and whether Institutional Review Board (IRB) approvals (or an equivalent approval/review based on the requirements of your country or institution) were obtained?
    \item[] Answer: \answerNA{} % Replace by \answerYes{}, \answerNo{}, or \answerNA{}.
    \item[] Justification: No human subjects research was conducted; thus, IRB approval was not required.

\item {\bf Declaration of LLM usage}
    \item[] Question: Does the paper describe the usage of LLMs if it is an important, original, or non-standard component of the core methods in this research? Note that if the LLM is used only for writing, editing, or formatting purposes and does not impact the core methodology, scientific rigorousness, or originality of the research, declaration is not required.
    %this research? 
    \item[] Answer: \answerNA{}
    \item[] Justification: Large language models were not used as part of the core methodology or experimental components of this research. Any use of LLMs was limited to minor editing and formatting support and does not impact the scientific contributions.

\end{enumerate}

\end{document}